\documentclass[11pt,a4paper]{article}
\usepackage[hyperref]{emnlp2020}
\usepackage{times}
\usepackage{latexsym}

\usepackage{enumitem}

\usepackage{microtype}

\aclfinalcopy 

\title{Does Social Support Expressed in Post Titles Elicit Comments in Online Substance Use Recovery Forums?}

\author{Anietie Andy\textsuperscript{1}, Sharath Chandra Guntuku\textsuperscript{2}\\
\textsuperscript{1}Penn Medicine 
\textsuperscript{2}Computer and Information Science \\
University of Pennsylvania \\
        {\tt \{anietie.andy@pennmedicine,sharathg@cis\}.upenn.edu}}       

\date{}

\begin{document}
\maketitle
\begin{abstract}
Individuals recovering from substance use often seek social support (emotional and informational) on online recovery forums, where they can both write and comment on posts, expressing their struggles and successes. A common challenge in these forums is that certain posts (some of which may be support seeking) receive no comments. In this work, we use data from two Reddit substance recovery forums: \textit{/r/Leaves} and \textit{/r/OpiatesRecovery}, to determine the relationship between the social supports expressed in the titles of posts and the number of comments they receive. 
We show that the types of social support expressed in post titles that elicit comments vary from one substance use recovery forum to the other. 
\end{abstract}

\section{Introduction} \label{introduction}

In the United States (US), substance use disorder (SUD) is one of the main causes of premature death \cite{johnston2003monitoring, schulte2013substance}. Social media is often used by people with SUD to seek support \cite{maclean2015forum77}. Online substance use recovery forums such as the Reddit forum, \textit{/r/Leaves} - which focuses on discussions around quitting marijuana, 
 provide a ``safe space" where members - some of whom may be struggling with substance use, can freely 
communicate and seek help from other members. 

Prior work showed that 10\% of support-seeking messages on a forum focused on discussions around cancer received no comments \cite{wang2015eliciting,yang2019seekers}. 
We find that support-seeking posts (across a 3 months time period) from  \textit{/r/Leaves} follow a similar trend with 11\% of posts receiving no comments, thereby leaving some users seeking support without adequate support. 

The following social supports: emotional and informational are crucial in online forums focused on discussions around health and well-being \cite{wang2012stay,yang2017commitment}, where similar to prior work \cite{wang2012stay}, emotional support sought in posts seek affirmation, encouragement, and compassion, while informational support sought in posts seek information or advice. 

Reddit posts are made up of two sections: the \textit{title} - which briefly describes the post, and the \textit{selftext} - which describes the post in more detail. 
\newcite{glenski2017predicting} demonstrated that on Reddit, user interactions such as voting and commenting on posts is guided by the title of the post (post-titles); thereby implying that readers utilize linguistic cues in the post-titles on Reddit to decide whether to respond to posts. 
Hence, in this work, we focus on analyzing the post-titles in two Reddit substance use recovery forums \textit{/r/Leaves} and \textit{/r/OpiatesRecovery}. We measure the extent to which these  post-titles seek emotional support and informational support and analyze their relationship with the number of comments they receive. 
We hypothesize that since the interests of users who
belong to different forums focused on similar discussions differs \newcite{tran2016characterizing}, the type of social support (expressed in post-titles) that elicits comments may differ from one online substance use recovery forum to the other.

Understanding the relationship between the social supports (emotional and informational) sought in post-titles published in substance use recovery forums and the number of comments these posts receive is important partly because users join online health forums when going through a health care event such as cancer \cite{wen2012understanding,yang2019seekers} or recovering from substance use \cite{maclean2015forum77}; therefore, posts not receiving comments in these forums means that some users are not getting the necessary support they need.




\section{Related Work}
Some prior work focused on non-health related forums while others focused on health related forums.

In non-health related forums, \newcite{althoff2014ask} 
studied a Reddit forum to determine the social and linguistic factors associated with posts in the forum that elicit responses. 
\newcite{tran2016characterizing} explored several Reddit forums and determined that different forums use different language styles and there was a correlation between a forums language style and the responses to comments. 
\newcite{jaech2015talking} examined the effect of language use in online forums on the reaction of members of the forums to comments. 
\newcite{hessel2017cats} studied if incorporating multimodal features attracted user attention \cite{horne2017identifying}.


In health related forums, \newcite{wang2012stay} examined the effect of different social supports users are exposed to in an online cancer forum and its effect on the duration of user membership in the forum.  
\newcite{wang2015eliciting} demonstrated that in an online 
cancer forum, members tend to respond with emotional support when users self-disclose negative information about themselves. 
\newcite{yang2019channel} studied communication in an online cancer forum and determined that members of the forum tend to disclose more negative information about themselves in public domains provided by the forums compared to the forums private domains. 
\newcite{yang2017commitment} examined the relationship between the kind of communication received by members of an online cancer forum and their commitment to the forum. 
\newcite{yang2019seekers} determined that over time, members of an online cancer forum change roles and that certain roles are predictors of prolonged commitment to the group. 
\newcite{chancellor2018norms} examined online weight loss forums and determined how support influences user behavior changes. \newcite{maclean2015forum77} analyzed different phases of opioid addiction in an online forum focused on discussions around opioid use recovery. 

Our work is different from prior work in that we analyze posts published in  online substance use recovery forums with the aim to determine if and how emotional and informational support sought in published posts in substance use recovery forums elicit comments from members of the forums. This study received exempt status from the University of Pennsylvania Institutional Review Board.

\section{Dataset} \label{dataset}

Our dataset consists of posts and meta-data from two active Reddit substance use recovery forums, \textit{/r/Leaves} and \textit{/r/OpiatesRecovery}, which have 147,000 members and 27,000 members, respectively, as of September 2020. \textit{/r/Leaves} is self-described as \textit{``a support and recovery community for practical discussions about how to quit pot, weed, cannabis, edibles, BHO, shatter, or whatever THC-related product you're using, and support in staying stopped"}. \textit{/r/OpiatesRecovery} is self described as \textit{``We are a group of people dedicated to helping each other kick the habit”}.
We chose \textit{/r/Leaves} and \textit{/r/OpiatesRecovery} because 
these forums are the Reddit substance use recovery forums focused on marijuana and opioid use, respectively, with the most number of users. 

Typically in these forums, a member writes a post (\textit{title} and \textit{selftext}) and other members respond to the post by either voting the post \textit{up} or \textit{down} or writing a comment. In this paper, we focus only on the comment responses to posts. Using Google’s BigQuery\footnote{\url{https://cloud.google.com/bigquery/}}, which is a data warehouse containing data from Reddit, 
we collected and processed all posts published in \textit{/r/Leaves} and \textit{/r/OpiatesRecovery} between December 2015 and August 2019 and collected the following data related to each post: the post-title, the user who published the post, the time the post was created, the comments the post received, and the number of comments the post received.
Table \ref{tab:dev} 
highlights the summary of our dataset. 

\begin{table}[h]
		\setlength{\abovecaptionskip}{5pt}
		\setlength{\belowcaptionskip}{-12pt}
		\small
		\centering
		\begin{tabular}{|l|l|l|}
			\hline
			\centering
			\textbf{Attribute} & \textbf{\textit{/r/Leaves}}&\textbf{\textit{/r/OpiatesRecovery}}\\ 
			\hline
		
			Number of unique users  & 18,100&4,374 \\
			Number of posts  &35,961 &9,900\\
			Number of comments  & 227,850&129,801\\
		
			\hline
			
		\end{tabular}
		
		\caption{Summary of the \textit{/r/Leaves} and \textit{/r/OpiatesRecovery} datasets}
		
		\label{tab:dev}
	\end{table}
	
\section{Social Support}
Several studies have shown the importance of the expression of emotional and informational social supports in forums focused on discussions around health \cite{wang2012stay,yang2017commitment,yang2019seekers}. 
Using a similar method from previous work by \newcite{wang2012stay}, we built two  models to determine 
how much emotional and informational support, respectively, is sought in post-titles in our dataset. 
We had 3 annotators - who were health care professionals with graduate degrees and familiar with substance use recovery, to rate a sample of 1,000 post-titles from our dataset, on (i) how much informational support each post-title sought and (ii) how much emotional support each post-title sought; where informational support post-titles provide/seek advice or information and emotional support post-titles seek encouragement, understanding, or affirmation \cite{wang2012stay}. Similar to prior work, \cite{wang2012stay}, the annotators rated these post-titles using a 7-point Likert scale (1 meant ``social support was not expressed in a post-title" and 7 meant ``social support was expressed a lot in a post-title"). To measure the reliability of the annotators, we used intra-class correlation (ICC) \cite{bartko1966intraclass}, which measures annotator reliability when each post-title is rated by different groups of annotators; 
the ICC for informational support sought and emotional support sought were 0.95 and 0.93, respectively. For each post-title, the annotator ratings were averaged, hence each post-title had a score that ranged between 1 and 7 which indicated how much informational and emotional support was sought. 
	
\subsection{Features}

We extracted several language features from the annotated post-titles.

Using Linguistic Inquiry and Word Count (LIWC) \cite{pennebaker2015development} - a dictionary comprising different psycho-linguistic categories, we selected the following LIWC categories relevant to informational and emotional support \cite{wang2012stay}: "positive emotion", "negative emotion", "she/he", "you", "we",  "i",impersonal pronoun", auxiliary" "verb", "past", "present", "future","religion", "death" "they", "cognitive mechanism", "biological processes", "time".

Similar to \newcite{wang2012stay}, from each post-title, we extract the number of sentences, the number of words in each sentence, the number of sentences that contain negation words/phrases such as "not", and the number of sentences phrased as questions.

We also extracted the number of specific parts-of-speech and the number of strong subjectivity words such as "affirmation" and weak subjectivity words such as "abandoned" \cite{wilson2005recognizing, wang2012stay}.


Post-titles seeking advice or involving requests were identified and counted \cite{wang2012stay} by (i) selecting sentences that began with the word "you" and followed by a Modal verb such as "may" and (ii) selecting sentences that began with the word "please" and followed by a verb.


We collected names of medicines from the Food and Drug Administration website website \footnote{\url{https://www.fda.gov/Drugs/}}; also a comprehensive list of nicknames for drugs was compiled \cite{wang2012stay}; we counted the number of drug names in each post-title.
 
 
 We trained a model of 20 Latent Dirichlet Allocation (LDA) \cite{blei2003latent} topics 
 from 45,000 randomly selected post-titles from our dataset. Two physicians, who are familiar with substance use recovery, manually assigned a label to each of the topics, as shown in Table \ref{tab:features_topics}.

\begin{table*}[!t]
\small
	\caption{Summary of LDA topic themes and top 5  highly correlated words associated with each topic}
		\centering
		\begin{tabular}{|l|l|}
			\hline
			\centering
				
			\textbf{LDA topic themes}& \textbf{Highly correlated words}\\
			
            \hline
          
           Time sober& months, weeks, sober, clean, year\\
            Wanting reasons to stop & stop, can't, high, anymore, friends\\
            Feelings: anxious and depressed&feeling, depressed, tired, high, anxious\\
          Time sober before relapse& days, sober, clean, hours, free\\
           Withdrawal symptoms and addiction& withdrawal, drug, test,addiction,job, \\
          
          Cravings and relapse  & night, strong, cravings, weekend, relapsed\\
          Advice for quitting  &quitting, advice, tips,support, benefits\\
          Ready to quit  & time, quit, finally, hard, stop\\
          Making the decision to quit  & made, make, life, things, friends \\
           Feelings of quitting &anxiety, quitting, depression, deal,pains\\
          mood and feelings  & good, today, bad, high,start\\
            Years of using&years, daily, heavy, user, habit\\
           Reason for quitting& quitting, brain, back, motivation, fog\\
        Impact of addiction    &life, addiction, love, hate, relationship \\
       Quitting     &quit, ago, months,stop, haven't\\
         Side effects  &dreams, sleep, night, nightmares, insomnia\\
          Struggling with relapse  &day, end, relapse, struggling, thoughts\\
         Time to start sobriety   & day, today, start,journey, sobriety\\
          Thinking of quitting  &cold, turkey, thinking, cut, habit\\
            Sharing stories  & story, thought, wanted, share, addiction\\
			\hline
			
		\end{tabular}
	
		\label{tab:features_topics}
	\end{table*}

\begin{table}[h]
		\setlength{\abovecaptionskip}{5pt}
		\setlength{\belowcaptionskip}{-12pt}
		\small
		\centering
		\begin{tabular}{|l|l|}
			\hline
			\centering
			\textbf{Emotional Support} & \textbf{Informational Support}\\ 
			\hline
			
			Thinking of quitting- (0.76) & Share stories -  (0.69)\\
			Word length - (0.033) & Thinking of quitting-(0.046) \\
			Sharing stories - (0.032)  &  Word length - (0.044)\\
			Noun - (0.021) & Noun (0.035) \\
			Strong subjectivity - (0.016) & Strong subjectivity -  (0.028) \\
			
			\hline
			
		\end{tabular}
		
		\caption{Top 5 most important features as ranked by the Random forest model}
		
		\label{tab:feature_ablation}
	\end{table}

\subsection{Social Support Prediction Model}

We built two Random Forest models that each output a numerical value that indicates how much emotional support and informational support is expressed in a post-title based on the annotations. We experimented with SVM, logistic regression, and Random Forest; Random Forest performed better. 
We randomly partitioned the annotated posts into a training set (80\%), a validation set (10\%), and a test set (10\%). 
We used the validation set to evaluate the performance of the models; 
when the performance on the validation set was satisfactory, the models were then evaluated on the test set. Similar to \newcite{wang2012stay}, Pearson's correlation was used to measure the correlation of the models with the annotated data.

These models correlated with the average annotator ratings with 
Pearson's correlation r = 0.46 and r = 0.51 for seeking emotional support and informational support, respectively. We then applied these models to the posts in our dataset. Table \ref{tab:feature_ablation} shows the 5 most important features, as ranked by the Random forest model.

\section{Does social support expressed in post-titles elicit comments?}

A challenge in substance use recovery forums is that several support-seeking posts do not receive any comments, thereby leaving some posters potentially without adequate support. In this section, we aim to determine if emotional and informational support expressed in post-titles elicit comments in \textit{/r/Leaves} and \textit{/r/OpiatesRecovery}. For our analysis, we select posts by users with 5 or more published posts in our dataset i.e. 12,960 posts published by 1,285 users from \textit{/r/Leaves} and 
4,055 posts published by 335 users from \textit{/r/OpiatesRecovery}. For each user, we calculate the average number of comments they received for all their posts. We also calculated the average extent of emotional and informational support scores for posts published by each user.  
We correlate the mean extent of emotional and informational support, respectively with the mean number of comments received per user. 

\begin{table}[h]
		\setlength{\abovecaptionskip}{5pt}
		\setlength{\belowcaptionskip}{-12pt}
		\small
		\centering
		\begin{tabular}{|l|l|}
			\hline
			\centering
			\textbf{Feature} & \textbf{Pearson $r$}\\ 
			\hline
			
			Emotional Support Sought &  0.17\\
			Informational Support Sought & 0.22\\
			
			\hline
			
		\end{tabular}
		
		\caption{Correlation between social support sought and number of comments: \textit{/r/Leaves}. $p<0.001$. Number of users = 1,285} 
		
		\label{tab:emo_info_corr}
	\end{table}

\begin{table}[h]
		\setlength{\abovecaptionskip}{5pt}
		\setlength{\belowcaptionskip}{-12pt}
		\small
		\centering
		\begin{tabular}{|l|l|}
			\hline
			\centering
			\textbf{Feature} & \textbf{Pearson $r$}\\ 
			\hline
			
			Emotional Support Sought&  0.15\\
			Informational Support Sought & - 0.13\\
			
			\hline
			
		\end{tabular}

		\caption{Correlation between social support sought and number of comments:  \textit{/r/OpiatesRecovery}. $p<0.001$. Number of users = 335}
		
		\label{tab:emo_info_corr_opiate_recovery}
	\end{table}

\noindent
\textbf{Results and Discussion:}
 Tables \ref{tab:emo_info_corr} and \ref{tab:emo_info_corr_opiate_recovery} show the correlation between social support sought in post-titles and number of comments in \textit{/r/Leaves} and \textit{/r/OpiatesRecovery}, respectively. We observed that in \textit{/r/Leaves}, the average informational support sought by users in post-titles correlates more with the average number of comments received by these users compared to the average emotional support sought. Also, we observed that in \textit{/r/OpiatesRecovery}, the average emotional support sought had a positive correlation with the average number of comments received compared  the average informational support sought which had a negative correlation with the average number of comments received. 
 
  
  These findings can benefit members of substance use recovery forums; for example, users seeking to elicit comments to their posts published on \textit{/r/Leaves} may 
  use linguistic cues associated with higher informational support in their post-titles. Also, the findings from this work can benefit substance use recovery forum moderators; given that there is a negative correlation between the informational support sought in post-titles in \textit{/r/OpiatesRecovery}, this potentially means that some users seeking informational support in this forum are not receiving support. Hence moderators of the forum can come up with ways in which these informational support seeking posts receive comments; for example informational support seeking posts not receiving comments may be sent to moderators or users familiar with the support sought.\\
 
\noindent
 \textbf{Limitations and Future Work:} 
 
 
In our analysis, we focused on two subreddits -  \textit{/r/Leaves} and \textit{/r/OpiatesRecovery}, which are the forums focused on recovery from marijuana use and opiod use with the most number of members on Reddit. In the future, we would explore the relationship between social support expressed in posts and the responses (comments and votes) they receive, in other substance use recovery forums. While the results in this work indicate statistically significant
correlations, in the future, we would look at the affect of author tenure and Reddit karma (reputation) - all of which could potentially contribute to the response rates of posts. 
 

\section{Conclusion}
We built two models to measure the extent of informational and emotional social support expressed in post-titles in two substance use recovery. We used these models to show the social supports that elicit comments in these forums. 




\bibliographystyle{acl_natbib}
\bibliography{anthology,emnlp2020}


\end{document}